\documentclass[10pt,twocolumn]{article}

\usepackage{titlesec}
\titlespacing*{\section}
  {0pt}{1.6ex plus 0.5ex minus 0.3ex}{0.8ex plus 0.2ex}
\titlespacing*{\subsection}
  {0pt}{1.2ex plus 0.3ex minus 0.2ex}{0.6ex plus 0.2ex}

\usepackage{graphicx}
\usepackage{amsmath}
\usepackage{amssymb}
\usepackage{hyperref}
\usepackage{cite}
\usepackage{geometry}
\usepackage{graphicx}
\usepackage{subcaption}
\usepackage{booktabs} 

\title{LILAC: Long-sequence Incremental Low-latency Arbitrary Motion Stylization via Streaming VAE–Diffusion with Causal Decoding}

\author{
  Peng Ren$^{\dagger}$ \\
  \texttt{pren1@ucsc.edu}
  \and
  Hai Yang$^{\dagger}$ \\
  \texttt{yahai@ucsc.edu}
}

\date{}

\begin{document}
\sloppy

\maketitle
\footnotetext[1]{These authors contributed equally.}
\begin{abstract}
Generating long and stylized human motions in real time is critical for applications that demand continuous and responsive character control. Despite its importance, existing streaming approaches often operate directly in the raw motion space, leading to substantial computational overhead and making it difficult to maintain temporal stability. In contrast, latent-space VAE–Diffusion–based frameworks alleviate these issues and achieve high-quality stylization, but they are generally confined to offline processing. To bridge this gap, LILAC (Long-sequence Incremental Low-latency Arbitrary Motion Stylization via Streaming VAE–Diffusion with Causal Decoding) builds upon a recent high-performing offline framework for arbitrary motion stylization and extends it to an online setting through a latent-space streaming architecture with a sliding-window causal design and the injection of decoded motion features to ensure smooth motion transitions. This architecture enables long-sequence real-time arbitrary stylization without relying on future frames or modifying the diffusion model architecture, achieving a favorable balance between stylization quality and responsiveness as demonstrated by experiments on benchmark datasets. Supplementary video and examples are available at the project page: \href{https://pren1.github.io/lilac/}{https://pren1.github.io/lilac/}.
\end{abstract}

\section{Introduction}

Human motion stylization is essential for applications such as animation, virtual reality, and avatars. 
Although recent generative models based on variational autoencoders (VAEs) and diffusion achieve high-quality stylization, for example, MCM-LDM~\cite{song2024mcm}, they are designed for offline processing and cannot be directly applied in real-time scenarios, where motion must be generated continuously from streaming input.

In this work, we present a method that extends an offline VAE–Diffusion pipeline with arbitrary stylization capability~\cite{song2024mcm} into a streaming model for long, temporally consistent, real-time motion generation.

This approach makes three main contributions:
\begin{enumerate}
\item A latent-space architecture featuring a sliding-window causal design and a re-decoding/encoding mechanism that injects previously generated motion features into the latent representation, ensuring temporal continuity while requiring no access to future frames or modification to the diffusion model architecture.
\item Enables smooth and instantaneous transitions between arbitrary motion styles in a streaming setting, bringing offline style-conditioning methods into real-time operation.
\item Qualitative and quantitative evaluation on benchmark datasets, demonstrating the effectiveness of the proposed streaming framework compared to existing offline methods.
\end{enumerate}

\section{Related Work}

The task of \emph{motion generation} can be formulated as: given some input condition (e.g., an action label, text prompt, or past frames), synthesize a plausible human motion sequence. Early research focused on operating directly in the raw motion space (e.g., joint rotations and root trajectories) with recurrent or adversarial models. For example, Wang et al. proposed an LSTM-based generative
model with adversarial refinement, incorporating
contact information to produce realistic human
motions of infinite length~\cite{wang2018combining}.

Although such models demonstrated compelling results, working directly in the raw space is inefficient due to redundancy and noise in motion capture data, which leads to heavy computational overhead and limited temporal consistency. To overcome these limitations, Chen et al.\ introduced \emph{Motion Latent Diffusion (MLD)}, which encodes motion into a compact latent space via a VAE (see Figure~\ref{fig:vae_models}(a)) and applies conditional diffusion within that space~\cite{chen2023executing}. This design greatly improves scalability and reduces computational cost compared to raw-space diffusion, while maintaining generative quality. Following this transition, most subsequent research has focused on learning, controlling, and generating motion entirely within the latent space.

Within this latent domain, researchers have further explored \emph{motion stylization}, which aims to modify content motions with style features such as emotion or personality. Most existing works are restricted to an \emph{offline} setting, in which frames are assumed to depend simultaneously on past and future context, thereby requiring the entire motion sequence. Jang et al.\ proposed \emph{Motion Puzzle}, which enables arbitrary body-part motion style transfer while operating without explicit style annotations or paired training data~\cite{jang2022motionpuzzle}. Zhong et al.\ presented \emph{SMooDi}, a motion diffusion model that adapts a text-to-motion backbone with style guidance and a lightweight adaptor to generate diverse and realistic stylized motions~\cite{zhong2024smooodi}, and Song et al.\ introduced the \emph{MCM-LDM}, which disentangles and jointly conditions on content, trajectory, and style for arbitrary style transfer in latent space~\cite{song2024mcm}. While these methods deliver high-quality results, they rely on full-sequence inference and cannot support online generation.

Meanwhile, another line of research within the latent space focuses on real-time motion synthesis. \emph{T2M-GPT} combines a VQ-VAE for discrete representation learning with a GPT-based decoder for text-conditioned generation, though its tokenization causes noticeable information loss~\cite{t2mgpt}. \emph{MotionStreamer} further extends this idea to continuous latent representations, achieving text-conditioned streaming synthesis with causal predictions~\cite{xiao2025motionstreamer}. In parallel, \emph{MotionLCM} adopts latent consistency distillation to accelerate diffusion-based generation with only a few denoising steps~\cite{blb2024motionlcm}, and a subsequent variant, \emph{MotionPCM}, applies a phased consistency model for further acceleration~\cite{jiang2025motionpcm}.

While latent diffusion has advanced both stylization and real-time generation in separate veins, their intersection—real-time stylization within latent models—remains largely unexplored.
Some related efforts, such as \emph{Style-ERD}, perform online motion style transfer in a latent space using an encoder–recurrent–decoder architecture trained with an adversarial discriminator~\cite{tao2022styleerd}.
Our work lies at the intersection of \emph{stylization}, \emph{latent representation}, and \emph{streaming generation}, introducing a diffusion-based framework that maintains both arbitrary stylistic control and temporal consistency in real time.

\section{Method}
\subsection{Overview}

In this method, we convert an offline VAE–Diffusion pipeline~\cite{song2024mcm} into a real-time, long-sequence streaming framework. 
To clearly present our approach, we first define the problem setup in Section~\ref{sec:problem}. 
Section~\ref{sec:streaming} then introduces our streaming stylization framework, which employs a sliding-window strategy and latent re-decoding/encoding to ensure temporal consistency over arbitrarily long sequences. 
Finally, Section~\ref{sec:stylization} explains how diverse style embeddings are integrated into the diffusion model to achieve flexible and responsive stylization during streaming generation.

\subsection{Problem Definition}
\label{sec:problem}
In our setting, we apply stylization only to the local motion dynamics, while copying the original trajectory. This design ensures that user-intended paths remain unchanged, avoiding confusion in interactive applications such as VR, where altering trajectories could negatively impact user experience.

To formalize this idea, we represent the input motion sequence as 
$\mathbf{X} = \{x_{1}, x_{2}, \ldots, x_{T}\}$ of length $T$, 
where $x_{t} \in \mathbb{R}^{d}$ is the motion feature at time step $t$. 
Each motion frame is decomposed into trajectory and non-trajectory components:

\begin{gather}
x_{t} = \big[\tau(x_{t}), \; c(x_{t})\big], \\
\tau(x_{t}) \in \mathbb{R}^{3}, \quad c(x_{t}) \in \mathbb{R}^{d-3}.
\end{gather}

Here, $\tau(x_{t})$ denotes the root trajectory at time $t$, and $c(x_{t})$ represents the other local motion features.  

Given a style motion sequence $\mathbf{S} = \{s_{1}, \ldots, s_{K}\}$, the stylization task aims to generate a stylized motion sequence $\mathbf{Y} = \{y_{1}, \ldots, y_{T}\}$, where $y_{t} \in \mathbb{R}^{d}$, such that:

\begin{gather}
\tau(y_{t}) = \tau(x_{t}), \quad \forall t, \\
c(\mathbf{Y}) = \text{Stylize}\!\big(c(\mathbf{X}), \mathbf{S}\big).
\end{gather}

\begin{figure}[t]
    \centering
    \begin{minipage}[t]{0.45\columnwidth}
        \centering
        \includegraphics[width=\linewidth]{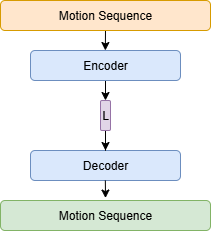}
        \caption*{(a) Baseline VAE}
    \end{minipage}
    \hfill
    \begin{minipage}[t]{0.45\columnwidth}
        \centering
        \includegraphics[width=\linewidth]{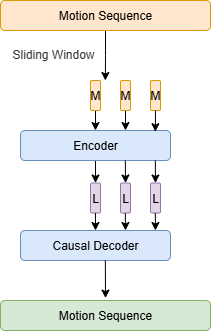}
        \caption*{(b) Proposed VAE}
    \end{minipage}
    \caption{
    Comparison between (a) the baseline VAE model and (b) the proposed VAE structure. 
    The baseline version directly encodes and reconstructs the motion sequence, 
    while the proposed variant introduces a sliding-window mechanism with a causal decoder 
    to process overlapping motion segments and improve temporal consistency, 
    particularly when the input is provided in a streaming manner.
    }
    \label{fig:vae_models}
\end{figure}

\begin{figure}[t]
    \centering
    \includegraphics[width=0.45\textwidth]{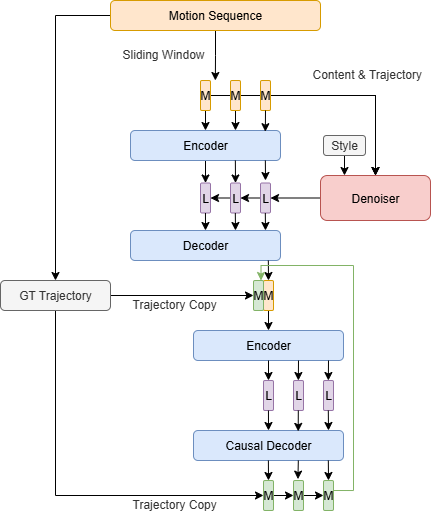}
    \caption{
    Proposed streaming stylization pipeline. 
    Motion segments are processed by an encoder–decoder with a style-conditioned denoiser. 
    The latent output of the diffusion model is re-decoded into motion features, 
    which are then concatenated with previous outputs, corrected through trajectory copy,
    re-encoded, and finally passed into a causal decoder for temporally consistent motion generation.
    }
    \label{fig:diffusion_model}
\end{figure}

\subsection{Streaming Stylization Framework}
\label{sec:streaming}
Offline frameworks are not directly applicable to streaming scenarios, since they require access to future data during inference.
A naïve adaptation—repeatedly applying the offline model to a sliding window of available data—results in severe temporal jitter, as diffusion models generate slightly different outputs at each run, causing discontinuities between adjacent segments. Moreover, transformer-based architectures cannot efficiently handle inputs that grow unbounded with sequence length, which further limits their direct application to long-sequence motion generation.

The offline framework~\cite{song2024mcm} consists of two components: (1) an encoder–decoder pair $(E, D)$ from a VAE, which maps motion sequences to and from the latent space, and (2) a diffusion model $\Phi$, which refines latent vectors conditioned on content, trajectory, and a pre-computed style embedding from MotionCLIP~\cite{tevet2022motionclip}. While $\Phi$ could in principle condition on previous outputs to generate continuous latents, we employ a sliding-window strategy that retrains both $\Phi$ and a causal variant of the VAE decoder, $\Psi$, to which the task of reconstructing temporally consistent motions is delegated.

As illustrated in Figure~\ref{fig:diffusion_model}, the input motion sequence $\mathbf{X}$ is segmented into overlapping windows $\mathbf{X}_{t:t+L}$ of length $L$, sampled with stride $\Delta$. Each window is first encoded into a latent representation by the VAE encoder $E$, stylized by the diffusion model $\Phi$ conditioned on content $\mathbf{c}_{t}$, trajectory $\tau_{t}$, and style embedding $\mathbf{s}$, and then decoded non-causally by $D$:

\begin{gather}
\mathbf{z}_{t} = E(\mathbf{X}_{t:t+L}) \\
\big[\tau_{t}, \; \mathbf{c}_{t}\big] = \mathbf{X}_{t:t+L} \\
\tilde{\mathbf{z}}_{t} = \Phi(\mathbf{z}_{t} \mid \mathbf{c}_{t}, \tau_{t}, \mathbf{s}) \\
\hat{\mathbf{Y}}^{\text{new}}_{t:t+L} = D(\tilde{\mathbf{z}}_{t})
\end{gather}

From the decoded output, only the newly generated portion is used and concatenated with the previously generated final results, after which the first three dimensions of $\mathbf{Y}^{\text{int}}$ are replaced with the trajectory from the input sequence:

\begin{gather}
\mathbf{Y}^{\text{int}}_{1:t+L} = 
\mathbf{Y}^{\text{prev}}_{1:t+L-\Delta} \,\Vert\, 
\hat{\mathbf{Y}}^{\text{new}}_{t+L-\Delta:t+L}, \\
\tau(\mathbf{Y}^{\text{int}}_{1:t+L}) = \tau(\mathbf{X}_{1:t+L}),
\end{gather}

To continue generation, the last $M$ frames of $\mathbf{Y}^{\text{int}}$ are re-encoded, fused with the most recent latent vector in a latent buffer $\mathcal{Z}$ using an exponential filter, and then appended back to $\mathcal{Z}$ while keeping only the latest $K$ elements.
\begin{gather}
\mathbf{z}_{\text{new}} = E(\mathbf{Y}^{\text{int}}_{t+L-M:t+L}) \\
\mathbf{z}_{\text{cur}} = \alpha \cdot \mathbf{z}_{\text{new}} + (1-\alpha)\cdot \mathbf{z}_{\text{recent}} \\
\mathcal{Z} \leftarrow \big(\mathcal{Z} \cup \{\mathbf{z}_{\text{cur}}\}\big)[-K:]
\end{gather}

The causal decoder $\Psi$ reconstructs the next stylized motion feature segment from the latent buffer $\mathcal{Z}$. The newly generated feature segment is concatenated with the previously accumulated motion features to form an updated feature sequence, with its trajectory component directly copied from the input motion $\tau(\mathbf{X})$, which is also stored as $\mathbf{Y}^{\text{prev}}$ for the next iteration.
All motion features are then mapped to joint coordinates, and only the most recent $\Delta$ joints are appended to the output sequence:
\begin{gather}
\hat{\mathbf{Y}}_{t+L-M:t+L} = \Psi(\mathcal{Z}) \\
\mathbf{Y}_{1:t+L}^{\text{new}} = \mathbf{Y}^{\text{prev}}_{1:t+L-\Delta} \,\Vert\, \hat{\mathbf{Y}}_{t+L-\Delta:t+L} \\
\tau(\mathbf{Y}_{1:t+L}^{\text{new}}) = \tau(\mathbf{X}_{1:t+L})\\
\mathbf{Y}^{\text{prev}}_{1:t+L} \leftarrow \mathbf{Y}_{1:t+L}^{\text{new}} \\
\mathbf{J}_{1:t+L} = f\!\left(\mathbf{Y}^{\text{new}}_{1:t+L}\right) \\
\mathbf{J}^{\text{final}} = \mathbf{J}^{\text{prev}} \,\Vert\, \mathbf{J}^{\text{new}}_{t+L-\Delta:t+L}
\end{gather} 

In theory, the proposed sliding-window design allows the model to process motion sequences of arbitrary length. Notice that if the available sequence length is shorter than the corresponding window size, all available frames are used in the corresponding step. This assumption holds for all of the preceding formulation.

\subsection{Stylization}
\label{sec:stylization}

\begin{figure}[t]
    \centering
    \includegraphics[width=0.45\textwidth]{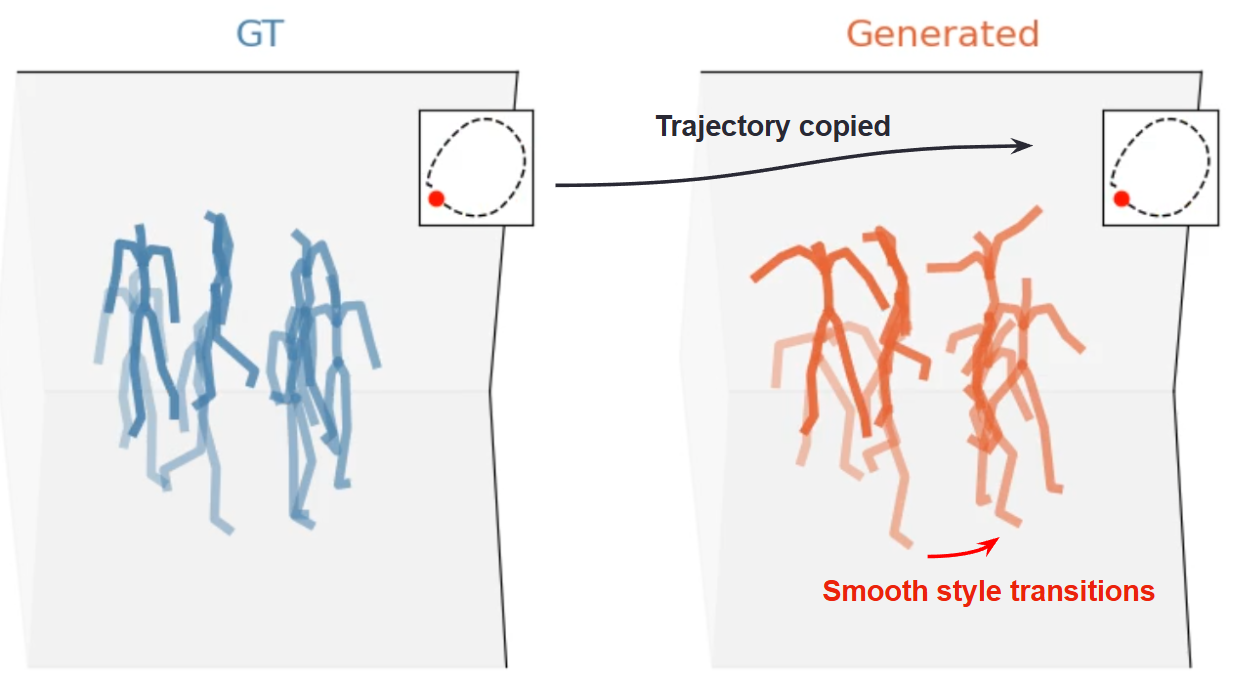}
    \caption{\textbf{Stylization example.} 
    The generated sequence (right) dynamically switches between multiple styles (\textit{old}, \textit{open-leg}, and \textit{hands-high}) in real time, while the ground-truth motion (left) — corresponding to the input reference — is shown for comparison. The color intensity indicates temporal progression, with darker poses corresponding to more recent frames. The small insets display the corresponding root trajectories, with a red dot indicating the end position of each trajectory.}
    \label{fig:stylization}
\end{figure}

As in the offline method, stylization is achieved by injecting the desired style embedding $\mathbf{s}$ into the diffusion model $\Phi$. In our online setting, this mechanism is naturally extended to allow style changes to appear almost immediately in the generated motion. At each stride $\Delta$, new latent features conditioned on updated styles are generated by the diffusion model and stitched together by the causal decoder $\Psi$, enabling continuous generation with responsive style transitions. An example of this dynamic style switching can be seen in Fig.~\ref{fig:stylization}.
\section{Experiments}

\subsection{Implementation Details}  
The VAE and diffusion model are retrained separately to enable robust motion stylization in streaming settings.

\smallskip
\noindent
\textbf{VAE configuration.}
As illustrated in Figure~\ref{fig:vae_models}(b), the VAE is trained by applying a sliding window to the input motion to produce latent sequences, which are then passed to the causal decoder. The encoder is kept frozen from the pretrained model, while only the causal decoder is retrained. During training, a random starting point within each sequence is sampled to ensure diverse temporal contexts. The latent dimension is set to $256 \times 7$. 
 
The VAE objective integrates several complementary losses to ensure accurate reconstruction and temporal coherence. 
Specifically, $M^{\text{traj}}$ and $M^{\text{feat}}$ denote the global trajectory and non-trajectory motion features, respectively, while $J$ represents the 3D joint positions, enforcing spatial reconstruction fidelity. Their reconstructed counterparts produced by the VAE decoder are denoted as $\hat{M}^{\text{traj}}$, $\hat{M}^{\text{feat}}$, and $\hat{J}$, respectively. In addition, smoothness terms are imposed on the temporal derivatives $\Delta M^{\text{traj}}$ and $\Delta J$ to penalize abrupt changes and enforce continuity in the generated motion. Note that $M^{\text{traj}}$ and $\Delta M^{\text{traj}}$ are included as part of the VAE training for regularization purposes, even though the final system later reuses the trajectory directly; this design choice is independent of the VAE formulation. The overall training objective is therefore given by:
\begin{gather}
\mathcal{L}_{\text{total}} = \lambda_{\text{traj}}\| \hat{M}^{\text{traj}} - M^{\text{traj}} \|_1 \nonumber \\
+ \lambda_{\text{feat}}\| \hat{M}^{\text{feat}} - M^{\text{feat}} \|_1 \nonumber
+ \lambda_{\text{joints}}\| \hat{J} - J \|_1 \nonumber \\
+ \lambda_{\text{smooth-traj}}\| \Delta \hat{M}^{\text{traj}} - \Delta M^{\text{traj}} \|_1 \nonumber \\
+ \lambda_{\text{smooth-joints}}\| \Delta \hat{J} - \Delta J \|_1 ,
\end{gather}
where the weights are fixed as 
$\lambda_{\text{traj}}=2.0$, 
$\lambda_{\text{feat}}=1.0$, 
$\lambda_{\text{joints}}=1.0$, 
$\lambda_{\text{smooth-traj}}=0.1$, and 
$\lambda_{\text{smooth-joints}}=0.1$.

\smallskip
\noindent
\textbf{Diffusion configuration.}
To improve robustness to temporal offsets introduced by the sliding-window setting, the diffusion model randomly selects the starting frame of each training sequence, rather than always beginning from the first frame as in the offline setting. The number of denoising steps was fixed at 10.

\smallskip
\noindent
\textbf{Streaming configuration.}
In the streaming pipeline, the window length is set to $L=60$, stride to $\Delta=4$, re-encoding length to $M=30$, latent buffer size to $K=30$, and blending weight to $\alpha=0.8$.
\subsection{Qualitative Results}
\begin{figure}[t]
    \centering
     \includegraphics[width=0.45\textwidth]{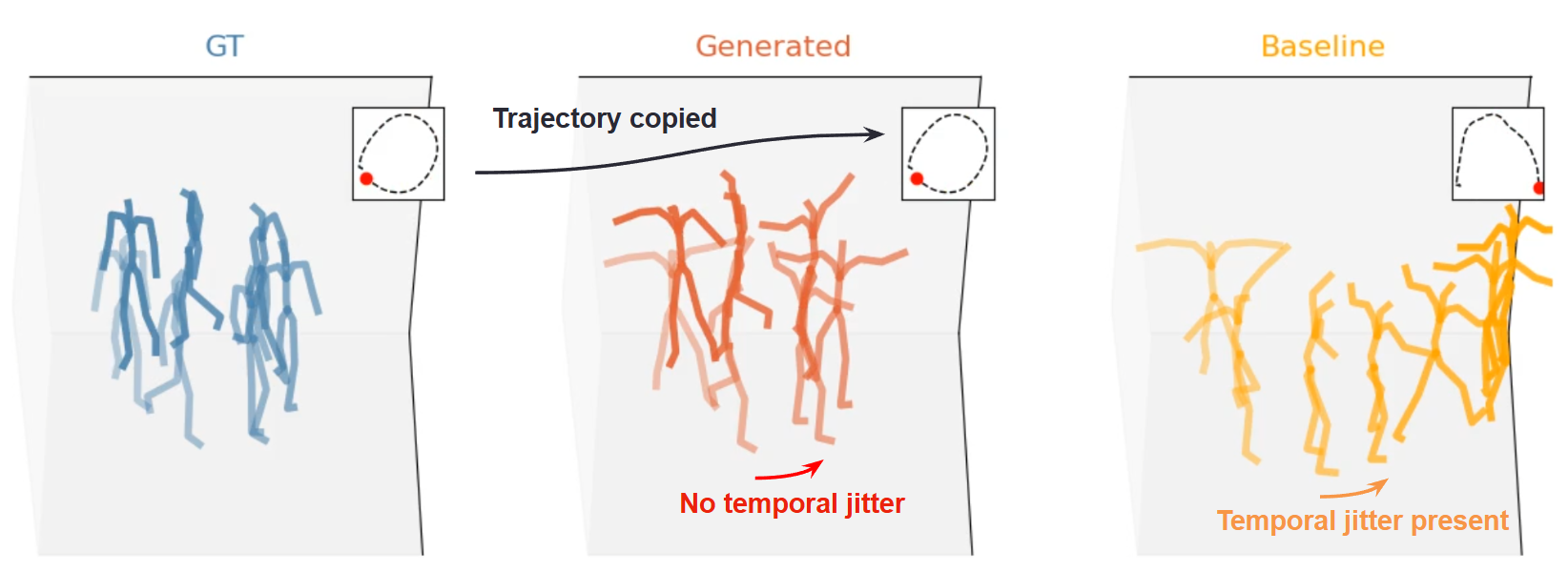}
    \caption{
    Qualitative comparison of long-sequence motion generation under the \textit{hands-high} stylization condition.
    The \textbf{ground-truth motion} (left) shows natural continuity. 
    The \textbf{proposed streaming model} (middle) produces smooth and temporally consistent stylized motion without jitter, 
    while the \textbf{offline baseline} (right) suffers from temporal jitter due to independent generation. Visualization conventions (e.g., color intensity and trajectory insets) follow those in Fig.~\ref{fig:stylization}. 
    }
    \label{fig:qualitative_jitter}
\end{figure}

Since the original paper only provides an offline framework, we construct a straightforward baseline to adapt it to the streaming setting, hereafter referred to as the \emph{streaming baseline}. Specifically, the initial $L$ frames are fed into the offline model to generate an output sequence of length $L$. At each subsequent step, the input window is shifted forward by $\Delta$ frames, and a new sequence of length $L$ is generated. The last $\Delta$ frames of this newly generated sequence are then concatenated to the previously produced results, and the process repeats until the entire motion sequence is completed.

As illustrated in Fig.~\ref{fig:qualitative_jitter}, the proposed streaming framework generates smooth and temporally consistent motion for the \textit{hands-high} stylization case, while the \emph{streaming baseline} shows segment discontinuities and temporal jitter. These results demonstrate the effectiveness of our approach in maintaining temporal coherence. For more detailed visual comparisons, including the long-sequence evaluation results, please refer to the supplementary video and additional examples available on the project page (link provided in the abstract).


\subsection{Quantitative Results}
\begin{table}[t]
\centering
\resizebox{\linewidth}{!}{%
\begin{tabular}{lrrrrr}
\toprule
Methods & FMD ↓ & CRA ↑ & SRA ↑ & Total Jitter ↓ \\
\midrule
Real Motions         &   --   & 0.99 & 1.00 & -- \\
\midrule
1DConv+AdaIN~\cite{huang2017adain}         &   42.68   & 0.31 & 0.57 & -- \\
STGCN+AdaIN~\cite{yan2018stgcn}          &   129.44   & \textbf{0.60} & 0.18 & -- \\
Motion Puzzle~\cite{jang2022motionpuzzle}        &   113.31   & 0.26 & 0.46 & -- \\
\midrule
Offline (Original Paper)~\cite{song2024mcm} & \textbf{27.69}  & 0.36 & 0.58 & \textbf{0.0089} \\
Streaming Baseline (Naïve)  & 39.94 & 0.27 & \textbf{0.59} & 0.0245 \\
Online (4 steps)   & 34.34 & 0.28 & 0.57 & 0.0127 \\
\textbf{Online (Proposed, 10 steps)}  & 31.69 & 0.30 & 0.58 & 0.0139 \\
\midrule
Ablation (No Extra Encoder–Decoder)    & 67.52 & 0.20 & 0.52 & 0.0157 \\
Ablation (Non-Causal Decoder)          & 30.65 & 0.28 & 0.57 & 0.0176 \\
\bottomrule
\end{tabular}}
\caption{Quantitative comparison with baselines and our streaming variants.
\textbf{Bold numbers} highlight the strongest performance within each metric. 
Symbols “↑” and “↓” indicate whether higher or lower values are preferable, respectively. 
The 10-step configuration (\textit{Online, Proposed, 10 steps}) is adopted as our default
setting, offering the best trade-off between motion quality and real-time performance. Here, “step” denotes the number of denoising iterations in the diffusion process (e.g., 4 vs. 10 for speed–quality trade-off). Results for the offline model and prior baselines are reported from the original paper~\cite{song2024mcm}.}
\label{tab:quantitative}
\end{table}

We take the offline implementation from the original paper~\cite{song2024mcm} as an upper-bound reference and conduct our evaluation using its provided evaluation dataset and defined metrics—Fréchet Motion Distance (FMD), Content Recognition Accuracy (CRA), and Style Recognition Accuracy (SRA)—to directly assess the performance gap between the offline and online settings. For detailed definitions of these metrics, the reader is encouraged to consult the original paper~\cite{song2024mcm}. 

We additionally compute a temporal smoothness metric, \textit{Total Jitter}, 
which measures the average per-joint frame-to-frame velocity change (acceleration magnitude) in the generated motion sequences, reflecting the temporal continuity of synthesized motions.
The computation reuses the motion sequences generated for CRA and SRA evaluation, 
where each motion sequence $\mathbf{X}\in\mathbb{R}^{T\times 22\times 3}$ 
contains $T$ frames with 22 joints represented by 3D coordinates.
Formally, let $\mathbf{X}^{\text{CRA}}$ and $\mathbf{X}^{\text{SRA}}$ 
denote the collections of sequences used in CRA and SRA evaluation, respectively.
The overall jitter is defined as
\begin{equation}
\text{Total Jitter} =
\frac{
D(\mathbf{X}^{\text{CRA}}) + D(\mathbf{X}^{\text{SRA}})
}{
N(\mathbf{X}^{\text{CRA}}) + N(\mathbf{X}^{\text{SRA}}) },
\label{eq:jitter}
\end{equation}

where $D(\cdot)$ sums the Euclidean norms of second-order frame differences across all joints, and $N(\cdot)$ denotes the total number of joint pairs used in this calculation.
Lower jitter indicates smoother temporal transitions in the generated motion. For our proposed method, to ensure fair evaluation, we repeat the input motion to fill the sliding window from the beginning, but compute metrics only after the model reaches steady streaming, excluding the initial frames used solely for window filling. This setup also allows us to continuously evaluate the model over extended sequences, demonstrating its ability to maintain consistency in long-sequence generation. In addition, one anomalously short motion sample (003790-5.npy, only 6 frames, approximately 0.3 s at 20 Hz) is excluded, as it is shorter than a single sliding window and thus not applicable to the streaming setup.

As summarized in Table~\ref{tab:quantitative}, traditional motion stylization methods such as 1DConv+AdaIN~\cite{huang2017adain}, STGCN+AdaIN~\cite{yan2018stgcn}, and Motion Puzzle~\cite{jang2022motionpuzzle} exhibit much higher FMD values, indicating a large gap in generation fidelity compared with diffusion-based approaches.
In contrast, both the offline reference (from~\cite{song2024mcm}) and our proposed online variant achieve substantially lower FMD and more balanced CRA/SRA scores, demonstrating the advantage of latent diffusion-based stylization.
Compared with the offline model, the proposed online framework shows a slight decrease in CRA and a small increase in FMD and \textit{Total Jitter}, which is expected given its sliding-window inference.
Nevertheless, it maintains comparable SRA and produces coherent long-sequence motions in real time.
The naïve streaming baseline, on the other hand, suffers from the highest \textit{Total Jitter}, confirming that simple segment concatenation without temporal refinement introduces severe discontinuities.

Each 4-frame update ($\Delta = 4$) required about 0.11–0.15 seconds on an RTX 4090 GPU, corresponding to an effective rate above 20 FPS, which matches the 20 FPS sampling rate of the HumanML3D dataset~\cite{guo2022action2motion}. Although the streaming design introduces a small fixed latency of 4 frames, the system comfortably meets real-time performance for continuous motion generation. The diffusion process remains the primary computational bottleneck; reducing the number of denoising steps from 10 to 4 increases the frame rate substantially (above 80 Hz in our experiments), at the expense of a moderate performance drop in motion quality, as indicated by the comparison results in the table.

\subsection{Ablation Study}

To better understand the effect of each design choice in our streaming framework, we conduct two ablation experiments, each targeting a specific component of the proposed model. The first variant, \textit{No Extra Encoder–Decoder}, removes the intermediate re-decoding and encoding branch that follows the diffusion stage, feeding the latent code produced by the diffusion model directly into the causal decoder. The second variant, \textit{Non-Causal Decoder}, modifies the decoder so that, when processing each sliding window, it can access future frames within the window rather than relying strictly on causal masking.

As reported in Table~\ref{tab:quantitative}, removing the extra decoder–encoder branch causes a substantial degradation in FMD, CRA, and SRA, indicating that re-decoding and re-encoding, which inject previously generated results into the latent representation, help maintain motion quality and preserving content and style consistency. The non-causal variant achieves slightly better FMD but slightly lower CRA/SRA scores, and its jitter also increases, highlighting the role of the causal structure in imposing one-directional dependencies from historical to current motion latents and thereby maintaining the temporal consistency required for online decoding of causally dependent motion sequences. Overall, these results confirm that both the re-decoding/encoding branch and causal conditioning are essential for achieving stable and coherent streaming motion stylization.

\section{Conclusion}
We introduce a streaming arbitrary stylization framework that enables real-time, long-sequence motion generation by extending a VAE–diffusion model in the latent space beyond the limitations of prior approaches.
In contrast to methods restricted to offline inference or operating directly in raw motion space, our framework employs a sliding-window strategy with a causal decoder and a lightweight re-decoding/encoding step to preserve temporal consistency while supporting immediate style adaptation. Quantitative and qualitative evaluations confirm that our method generates stylized motions that remain smooth and consistent over long sequences, and the proposed framework can be readily extended to other VAE–Diffusion architectures with minimal structural adjustments.

In future work, we plan to refine the framework in three directions. First, we will investigate a streaming encoder that incrementally encodes incoming motion into the latent space, potentially reducing the need for overlapping windows.
Second, we aim to further improve temporal consistency by allowing the diffusion model to condition on previously generated latents. Finally, since the diffusion process remains the main computational bottleneck, exploring efficient sampling strategies such as MotionPCM~\cite{jiang2025motionpcm} is a promising direction to enhance runtime efficiency. These extensions would strengthen the practicality of the framework and support future deployment in real-time motion capture systems.

\section*{Acknowledgments}
We thank the authors of MCM-LDM~\cite{song2024mcm} for releasing their codebase, which served as the foundation for implementing and validating our streaming extension. This work was carried out as an independent research effort without external funding. The implementation and results are available upon request for academic discussion.
\bibliographystyle{unsrturl}
\bibliography{references}

\begin{thebibliography}{10}

\bibitem{song2024mcm}
Wenfeng Song, Xingliang Jin, Shuai Li, Chenglizhao Chen, Aimin Hao, Xia Hou, Ning Li, and Hong Qin.
\newblock Arbitrary motion style transfer with multi-condition motion latent diffusion model.
\newblock In {\em Proceedings of the IEEE/CVF Conference on Computer Vision and Pattern Recognition (CVPR)}, pages 821--830, June 2024.

\bibitem{wang2018combining}
Zhiyong Wang, Jinxiang Chai, and Shihong Xia.
\newblock Combining recurrent neural networks and adversarial training for human motion synthesis and control.
\newblock {\em IEEE Transactions on Visualization and Computer Graphics}, 27(1):14--28, 2021.
\newblock \href {https://doi.org/10.1109/TVCG.2019.2938520} {\path{doi:10.1109/TVCG.2019.2938520}}.

\bibitem{chen2023executing}
Xin Chen, Biao Jiang, Wen Liu, Zilong Huang, Bin Fu, Tao Chen, and Gang Yu.
\newblock Executing your commands via motion diffusion in latent space.
\newblock In {\em Proceedings of the IEEE/CVF Conference on Computer Vision and Pattern Recognition (CVPR)}, pages 18000--18010, June 2023.

\bibitem{jang2022motionpuzzle}
Deok-Kyeong Jang, Soomin Park, and Sung-Hee Lee.
\newblock Motion puzzle: Arbitrary motion style transfer by body part.
\newblock {\em ACM Trans. Graph.}, 41(3), June 2022.
\newblock \href {https://doi.org/10.1145/3516429} {\path{doi:10.1145/3516429}}.

\bibitem{zhong2024smooodi}
Lei Zhong, Yiming Xie, Varun Jampani, Deqing Sun, and Huaizu Jiang.
\newblock Smoodi: Stylized motion diffusion model.
\newblock In Ale{\v{s}} Leonardis, Elisa Ricci, Stefan Roth, Olga Russakovsky, Torsten Sattler, and G{\"u}l Varol, editors, {\em Computer Vision -- ECCV 2024}, pages 405--421, Cham, 2025. Springer Nature Switzerland.

\bibitem{t2mgpt}
Jianrong Zhang, Yangsong Zhang, Xiaodong Cun, Shaoli Huang, Yong Zhang, Hongwei Zhao, Hongtao Lu, and Xi~Shen.
\newblock T2m-gpt: Generating human motion from textual descriptions with discrete representations, 2023.
\newblock URL: \url{https://arxiv.org/abs/2301.06052}, \href {https://arxiv.org/abs/2301.06052} {\path{arXiv:2301.06052}}.

\bibitem{xiao2025motionstreamer}
Lixing Xiao, Shunlin Lu, Huaijin Pi, Ke~Fan, Liang Pan, Yueer Zhou, Ziyong Feng, Xiaowei Zhou, Sida Peng, and Jingbo Wang.
\newblock Motionstreamer: Streaming motion generation via diffusion-based autoregressive model in causal latent space, 2025.
\newblock URL: \url{https://arxiv.org/abs/2503.15451}, \href {https://arxiv.org/abs/2503.15451} {\path{arXiv:2503.15451}}.

\bibitem{blb2024motionlcm}
Wenxun Dai, Ling-Hao Chen, Jingbo Wang, Jinpeng Liu, Bo~Dai, and Yansong Tang.
\newblock Motionlcm: Real-time controllable motion generation via latent consistency model.
\newblock In Ale{\v{s}} Leonardis, Elisa Ricci, Stefan Roth, Olga Russakovsky, Torsten Sattler, and G{\"u}l Varol, editors, {\em Computer Vision -- ECCV 2024}, pages 390--408, Cham, 2025. Springer Nature Switzerland.

\bibitem{jiang2025motionpcm}
Lei Jiang, Ye~Wei, and Hao Ni.
\newblock Motionpcm: Real-time motion synthesis with phased consistency model, 2025.
\newblock URL: \url{https://arxiv.org/abs/2501.19083}, \href {https://arxiv.org/abs/2501.19083} {\path{arXiv:2501.19083}}.

\bibitem{tao2022styleerd}
Tianxin Tao, Xiaohang Zhan, Zhongquan Chen, and Michiel van~de Panne.
\newblock Style-erd: Responsive and coherent online motion style transfer.
\newblock In {\em Proceedings of the IEEE/CVF Conference on Computer Vision and Pattern Recognition (CVPR)}, pages 6593--6603, June 2022.

\bibitem{tevet2022motionclip}
Guy Tevet, Brian Gordon, Amir Hertz, Amit~H. Bermano, and Daniel Cohen-Or.
\newblock Motionclip: Exposing human motion generation to clip space.
\newblock In Shai Avidan, Gabriel Brostow, Moustapha Ciss{\'e}, Giovanni~Maria Farinella, and Tal Hassner, editors, {\em Computer Vision -- ECCV 2022}, pages 358--374, Cham, 2022. Springer Nature Switzerland.

\bibitem{huang2017adain}
Kfir Aberman, Yijia Weng, Dani Lischinski, Daniel Cohen-Or, and Baoquan Chen.
\newblock Unpaired motion style transfer from video to animation.
\newblock {\em ACM Trans. Graph.}, 39(4), August 2020.
\newblock \href {https://doi.org/10.1145/3386569.3392469} {\path{doi:10.1145/3386569.3392469}}.

\bibitem{yan2018stgcn}
Soomin Park, Deok-Kyeong Jang, and Sung-Hee Lee.
\newblock Diverse motion stylization for multiple style domains via spatial-temporal graph-based generative model.
\newblock {\em Proc. ACM Comput. Graph. Interact. Tech.}, 4(3), September 2021.
\newblock \href {https://doi.org/10.1145/3480145} {\path{doi:10.1145/3480145}}.

\bibitem{guo2022action2motion}
Chuan Guo, Xinxin Zuo, Sen Wang, Shihao Zou, Qingyao Sun, Annan Deng, Minglun Gong, and Li~Cheng.
\newblock Action2motion: Conditioned generation of 3d human motions.
\newblock In {\em Proceedings of the 28th ACM International Conference on Multimedia}, MM '20, page 2021–2029, New York, NY, USA, 2020. Association for Computing Machinery.
\newblock \href {https://doi.org/10.1145/3394171.3413635} {\path{doi:10.1145/3394171.3413635}}.

\end{thebibliography}

\end{document}